\documentclass[lettersize,journal]{IEEEtran}
\usepackage{amsmath,amsfonts}
\usepackage{amssymb}
\usepackage{algorithmic}
\usepackage{algorithm}
\usepackage{array}
\usepackage{booktabs,makecell,multirow}
\usepackage[caption=false,font=normalsize,labelfont=small,textfont=sf]{subfig}
\usepackage{textcomp}
\usepackage{stfloats}
\usepackage{url}
\usepackage{verbatim}
\usepackage{graphicx}
\usepackage{cite}
\usepackage{flushend}
\begin{document}

\title{DGP-Net: Dense Graph Prototype Network for Few-Shot SAR Target Recognition}

\author{Xiangyu Zhou, Qianru Wei, and Yuhui Zhang
}

\markboth{Journal of \LaTeX\ Class Files,~Vol.~14, No.~8, August~2021}%
{Shell \MakeLowercase{\textit{et al.}}: A Sample Article Using IEEEtran.cls for IEEE Journals}


\maketitle

\begin{abstract}
The inevitable feature deviation of synthetic aperture radar (SAR) image due to the special imaging principle (depression angle variation) leads to poor recognition accuracy, especially in few-shot learning (FSL). To deal with this problem, we propose a dense graph prototype network (DGP-Net) to eliminate the feature deviation by learning potential features, and classify by learning feature distribution. The role of the prototype in this model is to solve the problem of large distance between congeneric samples taken due to the contingency of single sampling in FSL, and enhance the robustness of the model. Experimental results on the MSTAR dataset show that the DGP-Net has good classification results for SAR images with different depression angles and the recognition accuracy of it is higher than typical FSL methods.
\end{abstract}


\section{Introduction}
\IEEEPARstart{S}{ynthetic} aperture radar (SAR) is an active coherent imaging system, which is not susceptible to the weather. Compared with traditional optical, infrared radar and other passive imaging systems, it has unique advantages in disaster monitoring resources exploration and military. However, image annotation is a very time-consuming and costly work for recognition. It is of practical significance to study how to recognize well in the case of limited labeled samples.

FSL, which can be thought as a matching problem, usually can be divided into four types: 1) data expansion of original few-shot data set by auxiliary information, such as TriNet\cite{ref2}; 2) pre-train the model on large-scale data, fine-tune the parameters of the top layers on the target few-shot data set\cite{ref3}; 3) make the model automatically learn some meta-knowledge, such as the initial super parameters, the structure, and optimizer, etc\cite{ref4}; 4) calculate the distance between samples by a given function to measure their similarity, such as Prototype Network\cite{ref5}, Matching Network\cite{ref6}, and Relation Network\cite{ref7} et al. These methods, which work by learning the representation of features, require the data to have sufficiently small intra-class distance and large inter-class distance. However, SAR image data does not have the above conditions, due to the low similarity of images under different depression angles of the same category. In other words, the feature distribution of the same category is relatively scattered. Thus, those methods mentioned above are not suitable.
\begin{figure}[!t]
\centering
\includegraphics[width=2.2in]{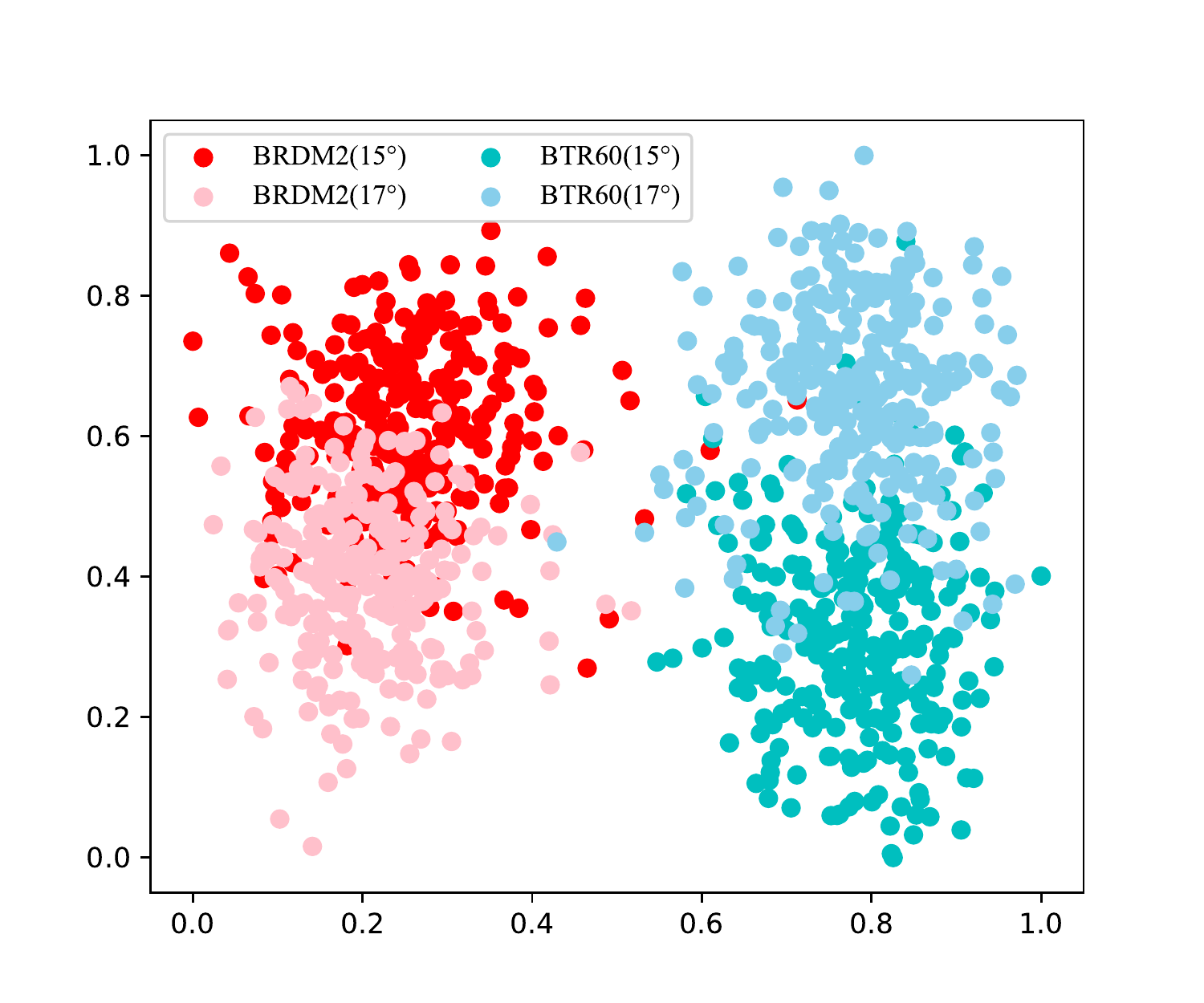}
\caption{Distribution of the features of two categories (BRDM2 and BTR60) SAR images with two depression angles ($15^\circ$ and $17^\circ$) extracted by CNN.}
\label{fig_1}
\end{figure}
\begin{figure}[!t]
\centering
\includegraphics[width=2.2in]{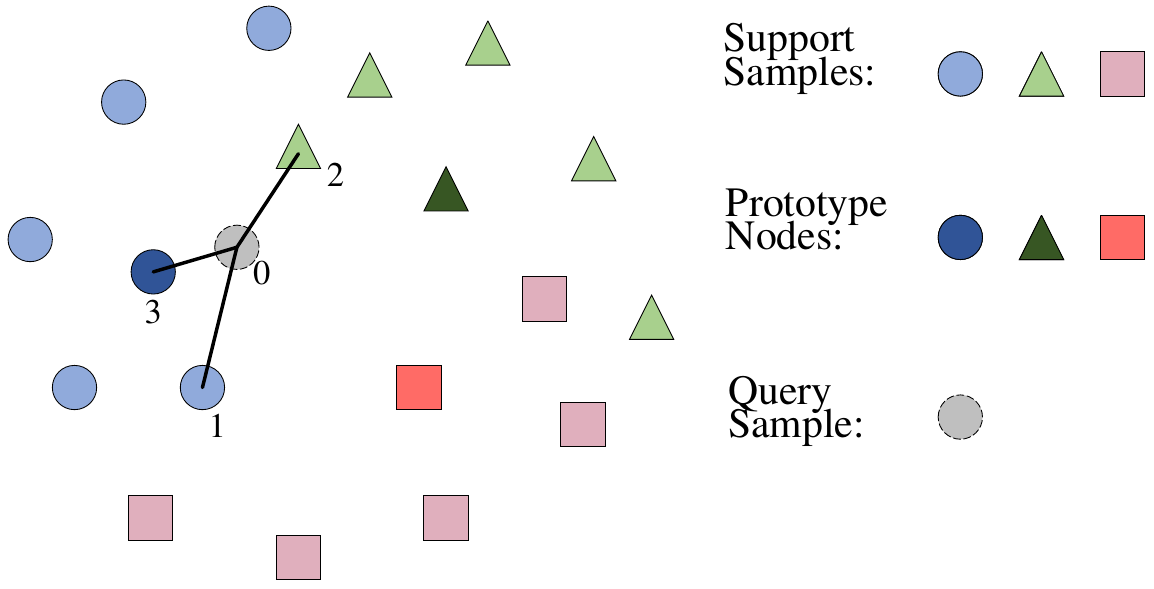}
\caption{The influence of adding prototype nodes on the classification effect.}
\label{fig_2}
\end{figure}

Our study finds that the feature deviation (i.e. the features of congeneric SAR images with different depression angles are offset in one direction) is to blame for this phenomenon, as shown in Fig. 1. Given this, we decide to use the graph convolutional network (GCN) to learn the distribution of features for recognition. And there have been some attempts to use GCN in FSL, such as \cite{ref8}, \cite{ref9}, and \cite{ref1}. GCN can flexibly describe the global manifold structure of sample distribution by the information diffusion mechanism. Information spreads in the network and becomes stable after several rounds of recursion, which is the final information expression\cite{ref11}. To eliminate the feature deviation, densely connected GCN is used to learn the potential features by information transmission. And the final distribution of features is learned for classification. 
\begin{figure*}[!t]
\centering
\includegraphics[width=7.2in]{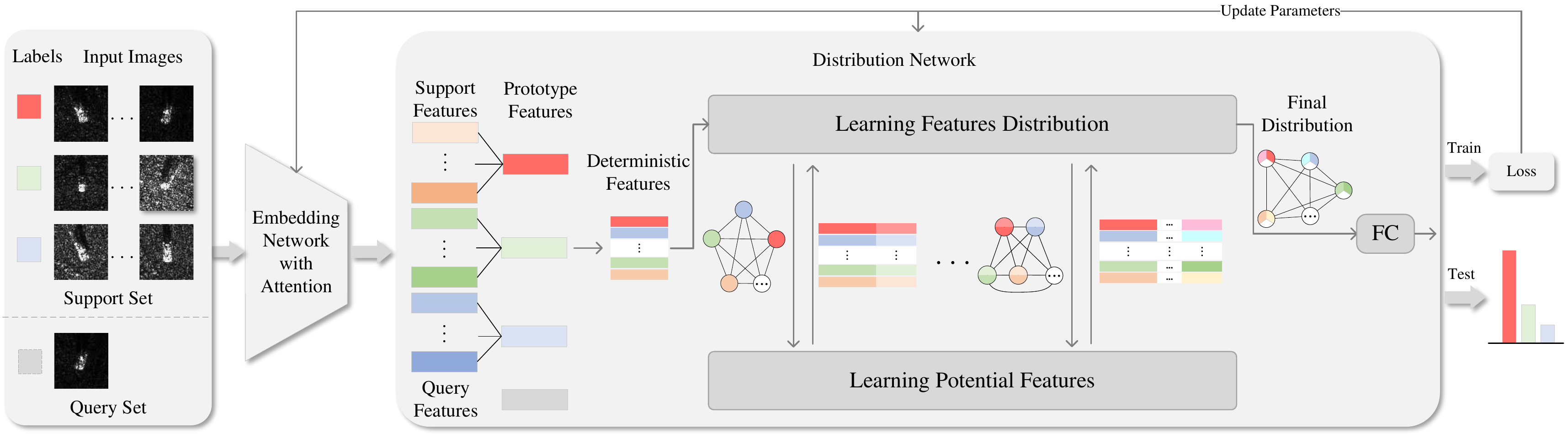}
\caption{Framework of the DGP-Net.}
\label{fig_3}
\end{figure*}

Due to the small sampling and large contingency of FSL, the features may be sampled at the edge of the distribution, resulting in inaccurate information transmission. Considering the robustness of the model, we add a prototype anchor point named "pro-point" for each category to gather congeneric samples and avoid misjudgment. As shown in Fig. 2, if the pro-points are not added, query sample 0 will be classified as $\Delta$ because it is closest to sample 2 and far from sample 1. For FSL, efficient feature extraction usually plays a crucial role in improving the effect \cite{ref10}. We add the attention mechanism to the embedding network to extract more useful features of images. In summary, we propose a dense graph prototype network (DGP-Net) for few-shot SAR target recognition. Our contributions are mainly reflected in the following aspects:
\begin{itemize}
\item{We propose DGP-Net, which effectively reduces the influence of different depression angles and achieves the best recognition effect so far. The design of the pro-point avoids the adverse effects of sampling at edges in FSL, and improves the robustness of the model.}
\item{This is the first study to explore the problem of feature deviation in few-shot SAR image recognition and solve it by learning the potential features and the distribution of features, which confirms the feasibility of using the GNN family to solve SAR FSL problems.}
\end{itemize}

To fully reflect the effectiveness of our method, we mix congeneric objects with different depression angles in MSTAR\cite{ref13} and develop a Hybrid-SAR dataset for experiments. And the DGP-Net achieves the best performance on it.

\section{Proposed Method}
DGP-Net is an end-to-end network architecture. As shown in Fig. 3, it consists of the embedding network for feature extraction and the distribution network for classification.

The training of the model is task-oriented. We set multiple groups of classification tasks, and conduct supervised training for each group of tasks. We update the parameters of both the embedding network and the distribution network by gradient descent. Through training, the model is expected to acquire the ability to distinguish different categories of images in the face of a new task with few samples. During the test, we fix the parameters of the entire DGP-Net and input labeled samples and the query sample into the model to obtain the recognition results.

\subsection{Data Organization and Definition of Task}
The data set is divided into training set $D_{tr}$ (including training subset $T_{tr}$ and test subset $T_{te}$) and test set $D_{te}$ (also including $T_{tr}$ and $T_{te}$), where the categories in $D_{tr}$ do not appear in $D_{te}$. The data of training and test task were sampled from $D_{tr}$ and $D_{te}$ respectively.
\begin{figure}[!t]
\centering
\includegraphics[width=2.5in]{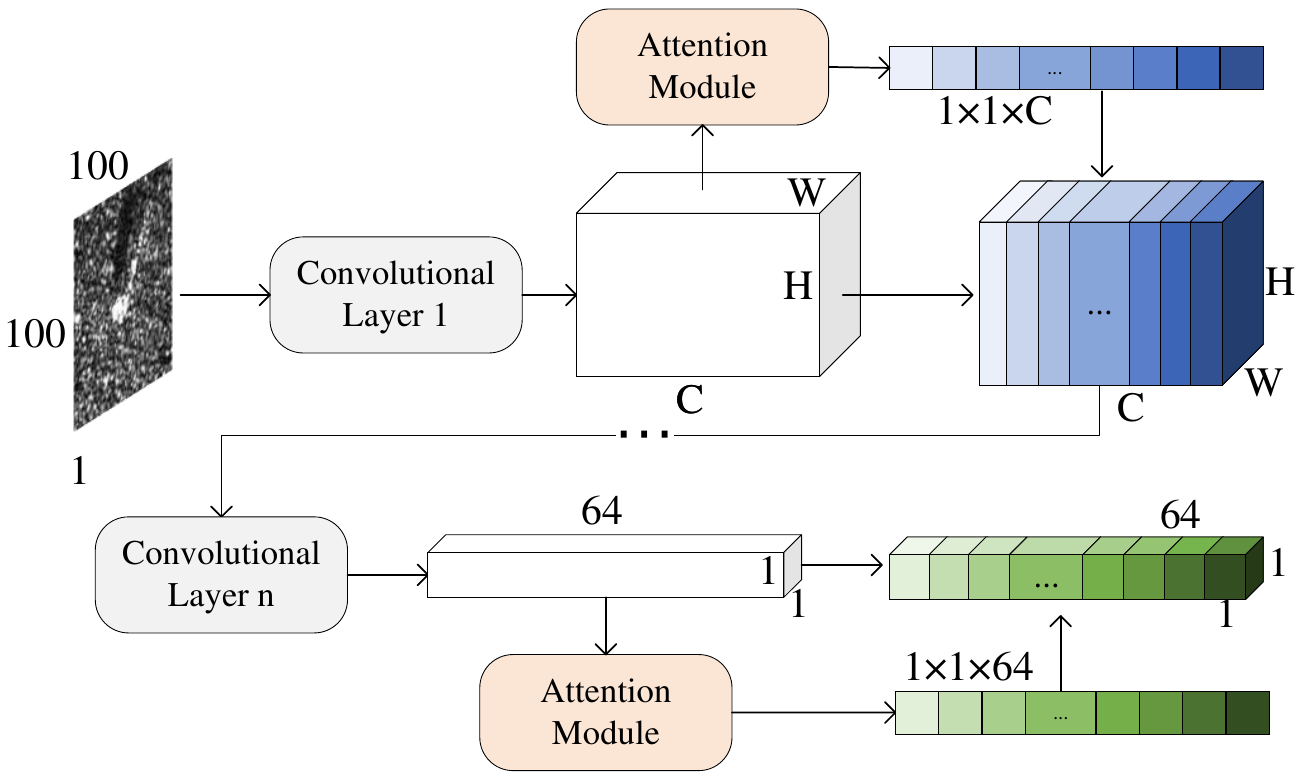}
\caption{Illustration of embedding network $f_{\phi}(x)$.}
\label{fig_4}
\end{figure}

For the N-way K-shot problem (i.e. the number of categories is $N$, and the number of support samples of each category is $K$), training data is sampled as follows: $N$ classes are randomly sampled in $D_{tr}$, and $K$ samples are randomly selected for each of the $N$ classes to construct $T_{tr}$ (i.e. there are $N \times K$ support samples in $T_{tr}$). A sample of one of these classes is randomly selected as query sample to construct $T_{te}$, and $T_{tr} \cap T_{te} = \varnothing $.

During the training of each task, all samples in $T_{tr}$ and $T_{te}$ are input into the model. The label of the query sample is used as the ground-truth for prediction. The training and test tasks share the same N-way K-shot problem, i.e., there are $N \times K$ labeled samples and a query sample input into the model when testing.

\subsection{Embedding Network}
The embedding network $f_{\phi}(x)$ is used for feature extraction, i.e., processing SAR images into one-dimensional vectors. As shown in Fig. 4, it contains a bunch of convolutional layers and several attention modules interspersed among them to learn the correlation between channels and change weights of each channel. With these modules, important information is enhanced and unimportant information is weakened, thus making the extracted features more useful. The embedding network outputs a vector of length 64.

The details of the attention module are as follows: firstly, we conduct a global average pooling of the feature map $M \in \mathbb{R}^{H \times W \times C}$ obtained by convolution:
\begin{equation}
\label{deqn_ex1a}
Z = \frac{1}{H \times W}\sum_{i=1}^H \sum_{i=1}^W m_c(i,j)
\end{equation}
where $Z \in \mathbb{R}^{C}$ is the channel-wise statistic, and $m_c \in M = [m_1,m_2,\cdots,m_C]$. Next, a fully-connected (FC) layer is used to learn the dependencies between channels. Finally, sigmoid is used to limit the value to the range of $[0,1]$ and obtain a one-dimensional vector $S$ with length of the channel number as the evaluation score:
\begin{equation}
\label{deqn_ex1a}
S = \sigma(\rho(\alpha Z))
\end{equation}
where $\sigma(\cdot)$ represents the sigmoid activation, $\rho(\cdot)$ represents the nonlinear activation function and $\alpha$ is a learned weight parameter. $S$ is multiplied with $M$ and used as the input feature map of the next convolutional layer:
\begin{equation}
\label{deqn_ex1a}
M^{\prime} = SM
\end{equation}

To construct the initial inputs of the distribution network, we extract feature vectors $ f_{\phi}(x_{i}) $ of all samples using the embedding network, and concatenate label information with the vectors using the following equation:
\begin{equation}
\label{deqn_ex1a}
X_{i} = (f_\phi(x_i) , l(y_i))
\end{equation}
where $X_{i}$ is the embedding vector of the sample $x_i$, $y_i$ is the label of $x_i$ and $l(y_{i})$ is the one-hot coding of $y_i$.

\subsection{Distribution Network}
This part is designed to obtain the distribution of features for classification. There are $N \times K + 1$ nodes, which are the embedding vectors of all samples in a task $\mathcal{T}$ input into the graph network. We associate $\mathcal{T}$ with a fully connected graph $G_\mathcal{T} = (V=\{v_i\},E=\{e_{i,j}\})$, where $e_{i,j}$ is the edge between node $v_i$ and node $v_j$, representing the similarity between $v_i$ and $v_j$, and $e_{i,j} = e_{j,i}$.

Firstly, calculate the prototype of congeneric features in $\mathcal{T}$ as the following equation:
\begin{equation}
\label{deqn_ex1a}
P_c = \frac{1}{|T_{tr}^c|}\sum_{x_i\in T_{tr}^c} X_i
\end{equation}
where $T_{tr}^c$ represents all samples of category $c$ in $T_{tr}$.

At this point, we have all the deterministic features as the prepared nodes for the next step:
\begin{equation}
\label{deqn_ex1a}
V = \{X_1 , \cdots , X_{N \times K }, P_1 , \cdots , P_N , X_{que}\}
\end{equation}
where $X_{que}$ is the embedding vector of the query sample $x_{que}$.

If the prototype information is added into the original nodes as partial features of the node, the updated node is formed by concatenating the original feature with the prototype of the category to which the original node belongs:
\begin{equation}
\label{deqn_ex1a}
X_{i} \gets (X_{i} , P_c), x_i\in T_{tr}^c
\end{equation}
In order to ensure that all nodes have the same length, $X_{que}$ is doubled:
\begin{equation}
\label{deqn_ex1a}
X_{que} \gets (X_{que} , X_{que})
\end{equation}
The prepared nodes are as follows:
\begin{equation}
\label{deqn_ex1a}
V = \{X_1 , \cdots , X_{N \times K }, X_{que}\}
\end{equation}

Next, learn the features distribution by constructing the adjacency matrix $A$. And the relation between every two nodes is expressed as:
\begin{equation}
\label{deqn_ex1a}
A_{i,j} = e_{i,j} = f_s(X_i , X_j)
\end{equation}
\begin{equation}
\label{deqn_ex1a}
f_s(X_i , X_j) = MLP(abs(X_i - X_j))
\end{equation}
where $f_s(X_i , X_j)$ represents the measure of similarity between node $v_i$ and $v_j$, which is obtained by training of multilayer perceptron (MLP), a simple artificial neural network. Its input is the absolute value of the difference of two feature vectors.

The potential feature learned by convolution is defined as:
\begin{equation}
\label{deqn_ex1a}
X^*_{i} = \rho(\alpha A X_{i})
\end{equation}
where $\rho$ represents the nonlinear activation function, $\alpha$ is a learned weight parameter, $A$ is the adjacency matrix, and $X_{i}$ represents the previous feature.

The network is densely connected to retain previous features. We concatenate previous features and potential features as new features:
\begin{equation}
\label{deqn_ex1a}
X^{\prime}_{i} = (X_{i} , X^*_{i})
\end{equation}
Thus, the length of the feature vector increases as the number of iterations increases. In general, the number of iterations is not set to more than 5 to ensure that the vector is not too long. If the network is formulated as non-densely connected, then:
\begin{equation}
\label{deqn_ex1a}
X^{\prime}_{i} = X^*_{i}
\end{equation}

After several iterations of learning features distribution and learning potential features, an adjacency matrix representing the final distribution can be obtained.

Finally, a full-connection layer is connected to output the possibility distribution of categories, which is used to calculate the cross-entropy loss:
\begin{equation}
\label{deqn_ex1a}
L = \sum_{i=1}^N y_{que}log(\hat{y}_{que})
\end{equation}
where $N$ is the number of categories, $y_{que}$ is the label of $x_{que}$, and $\hat{y}_{que}$ represents the categories possibility distribution of $x_{que}$.

In the training stage, update the parameters of the whole model by back-propagation:
\begin{equation}
\label{deqn_ex1a}
\Theta^{\prime} = \Theta - \eta \nabla L(\Theta)
\end{equation}
where $\Theta$ is the parameter set before update, $\eta$ is the learning rate, and $\nabla L(\Theta)$ represents the partial derivative of the loss with respect to $\Theta$.

\section{Experiment and Discussion}
\subsection{Data Set}
We evaluate our method on the MSTAR benchmark data set and select the data set sampled under standard operating condition (SOC) with two depression angles ($15^\circ$ and $17^\circ$), including 10 types of ground targets (T62, BTR60, ZSU234, BMP2, ZIL131, T72, BTR70, 2S1, BRDM2, and D7). And the Hybrid-SAR dataset is developed by mixing objects of the same category with different depression angles. The images in it are processed to size $100 \times 100$.

To ensure the consistency of the experiment, the categories in $D_{tr}$ and $D_{te}$ are fixed. For 5-way, the $D_{te}$ includes BTR60, BRDM2, T72, 2S1, and D7, the $D_{tr}$ consists of the other five categories. For 3-way, the $D_{te}$ includes BTR60, BRDM2, and T72, the $D_{tr}$ consists of the other seven categories.
\subsection{Experimental Settings}
\begin{figure}[!t]
\centering
\includegraphics[width=2.5in]{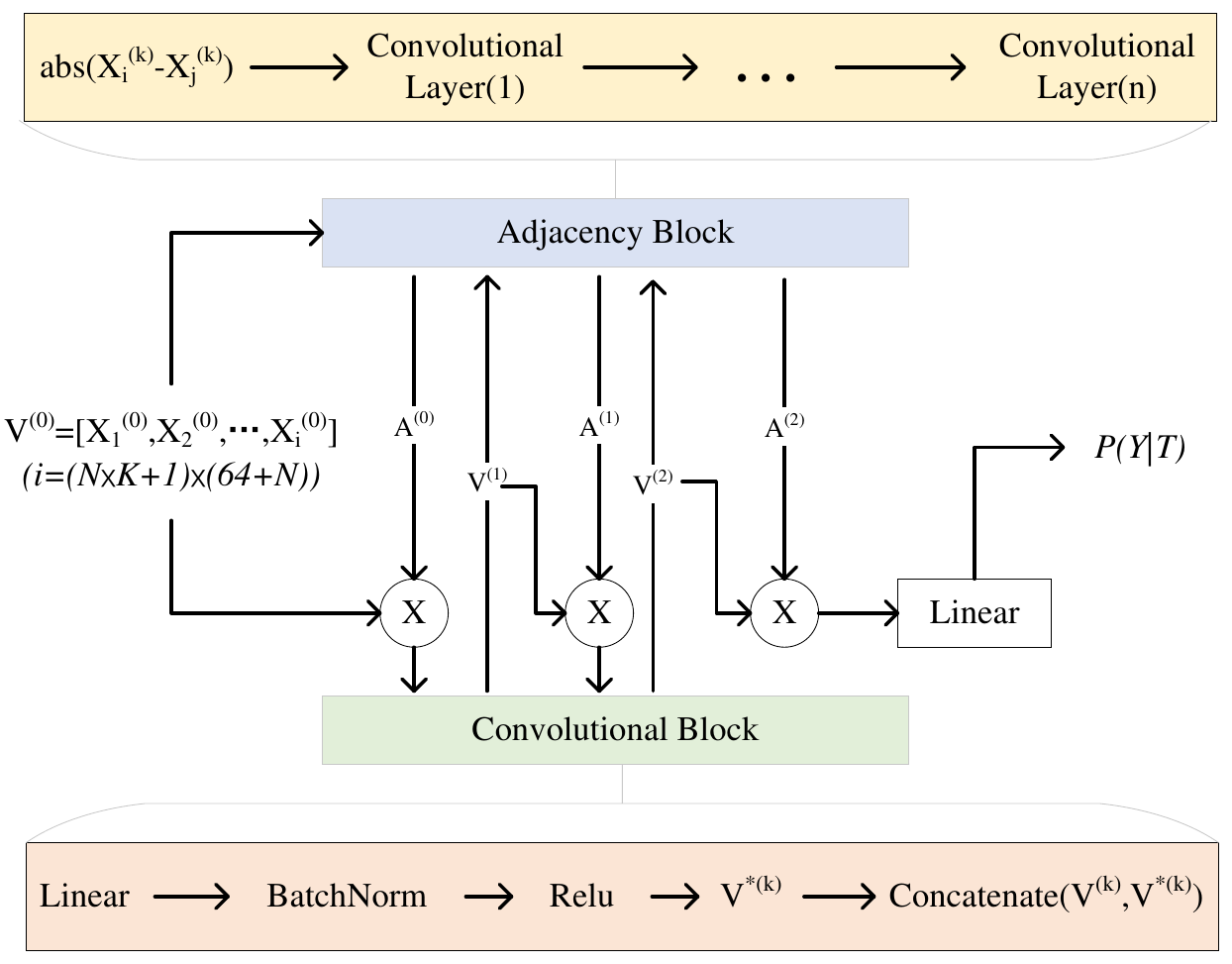}
\caption{Detailed structure of the distribution network.}
\label{fig_5}
\end{figure}
Combining with the SENet\cite{ref14}, we build the embedding network with attention mechanism. It consists of one convolutional layer: \{3$\times$3-conv(16 filters), BatchNorm, Relu\}, and three convolution blocks with the parameter $n$ (16, 32, 64). There are six basic blocks in each convolution block, each containing two convolutional layers: \{3$\times$3-conv(n filters), BatchNorm, Relu\}, \{3$\times$3-conv(n filters), BatchNorm\}, and an attention module: \{Avg\_pool, Linear, Relu, Sigmoid\}.

As shown in Fig. 5, the distribution network consists of the adjacency blocks and the convolution blocks. Convolution block is formed by a linear layer: \{Linear, BatchNorm, Relu\}, and a concatenate operation. We design different adjacency blocks for the 3-way and 5-way tasks. For 3-way, it is lighter and contains an operation to calculate Euclidean distance and three convolutional layers: \{1$\times$1-conv(64 filters), BatchNorm, Relu\}, \{1$\times$1-conv(32 filters), BatchNorm, Relu\}, \{1$\times$1-conv(1 filter)\}. And for 5-way, there are five convolutional layers: \{1$\times$1-conv(64 filters), BatchNorm, Relu\}, \{1$\times$1-conv(64 filters), BatchNorm, Relu\}, \{1$\times$1-conv(32 filters), BatchNorm, Relu\}, \{1$\times$1-conv(32 filters), BatchNorm, Relu\}, \{1$\times$1-conv(1 filter)\}.

Through experience and experimental comparison, we adopt the Adam optimizer with a learning rate of 0.001 for 3-way, and 0.01 for 5-way.

\subsection{Ablation Study}
In this section, some ablation experiments are performed on Hybrid-SAR dataset to verify the validity of the DGP-Net for weakening the feature deviation and clustering, and prove the effectiveness of dense connection and pro-point. 

\textit{1) Validity of the DGP-Net for weakening the feature deviation and clustering:} We use the t-distributed stochastic neighbor embedding (t-SNE)\cite{ref12} to visualize the distribution of feature vectors before and after the DGP-Net. As shown in Fig. 6 (a), the embedding vectors of congeneric samples are scattered. And there are obvious feature deviations between samples with different depression angles. The query sample labeled BTR60 is likely to be judged as BRDM2. It can be seen in Fig. 6 (d), that DGP-Net has eliminated the feature deviation, mixing the congeneric samples with different depression angles together and clustering congeneric samples away from different categories. And the query sample can be accurately identified as BTR60. Comparing Fig. 6 (b), Fig. 6 (c), and Fig. 6 (d), it can be seen that the elimination of feature deviation is the dense connection at work, and the pro-point plays an important role in clustering congeneric samples.
\begin{figure}[!t]
\centering
\subfloat[]{\includegraphics[width=1.7in]{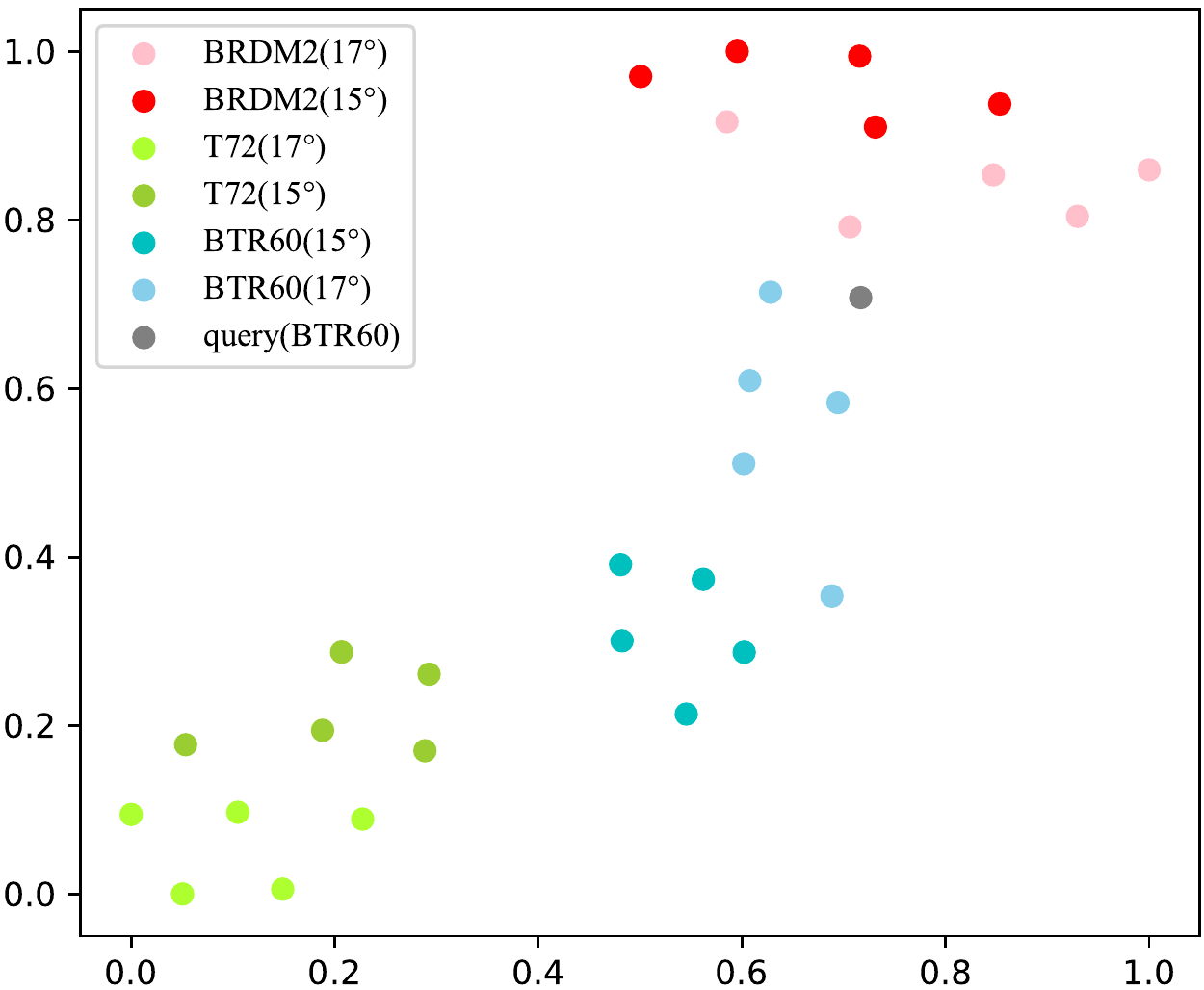}}
\hfil
\subfloat[]{\includegraphics[width=1.7in]{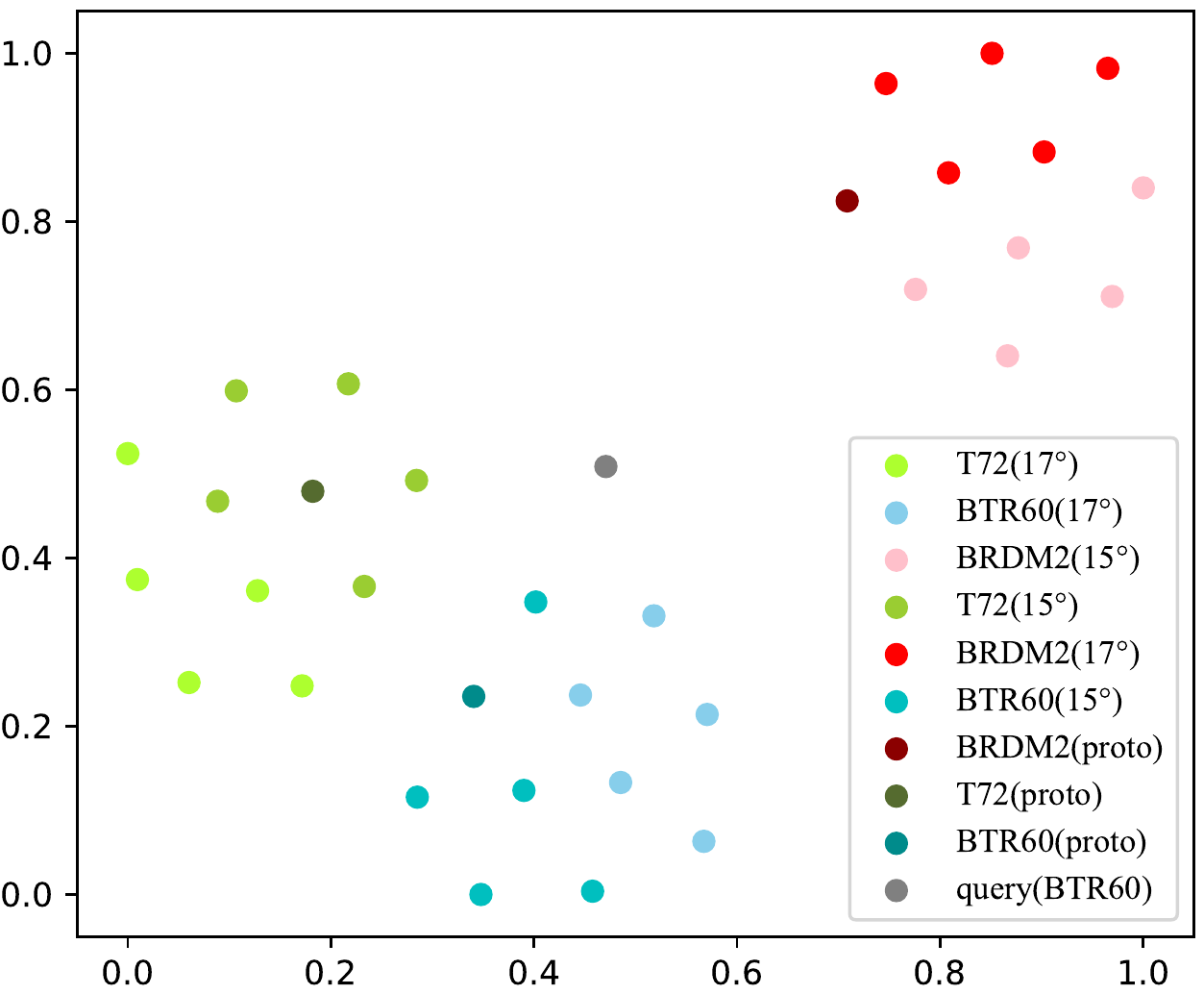}}
\hfil
\subfloat[]{\includegraphics[width=1.7in]{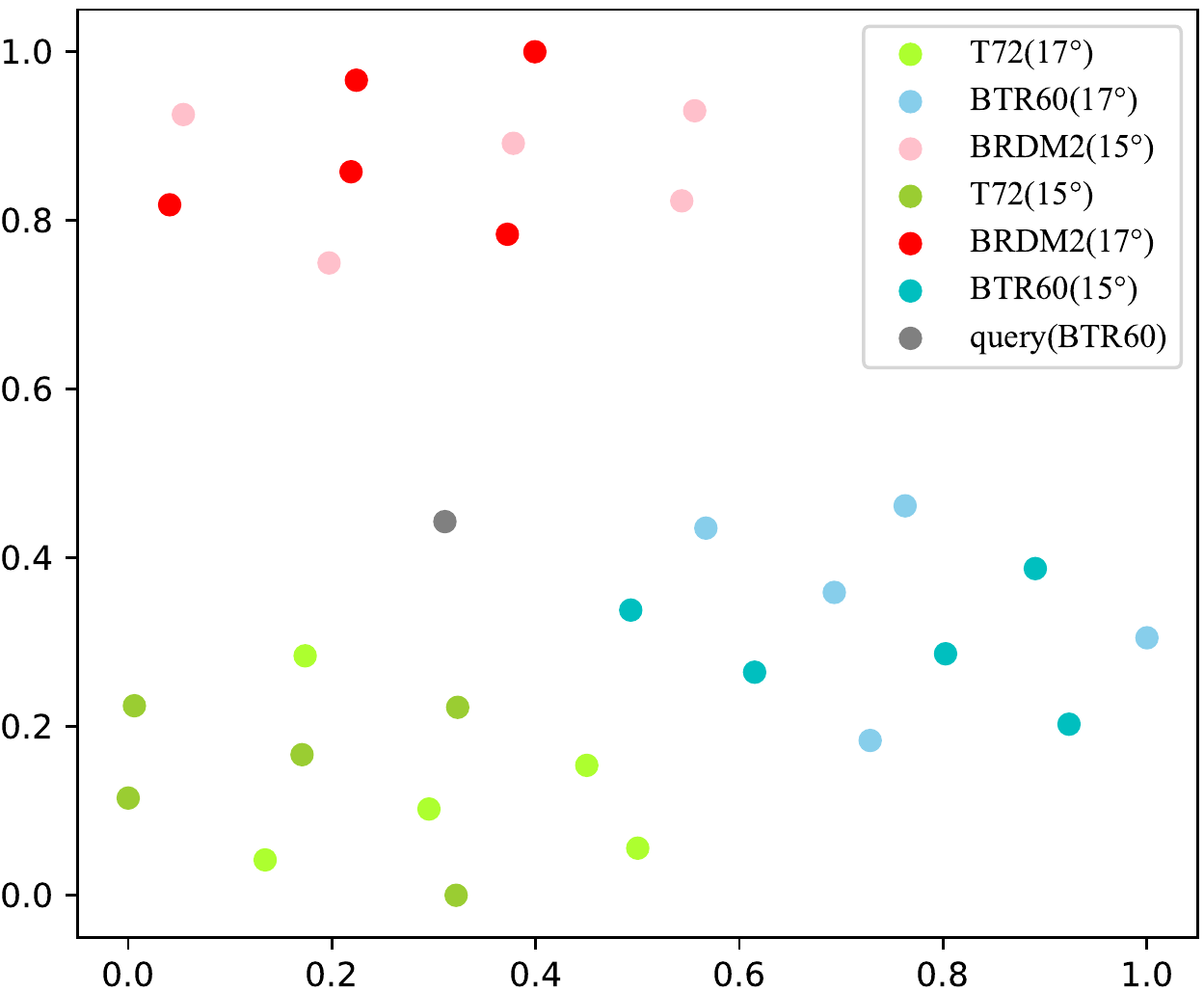}}
\hfil
\subfloat[]{\includegraphics[width=1.7in]{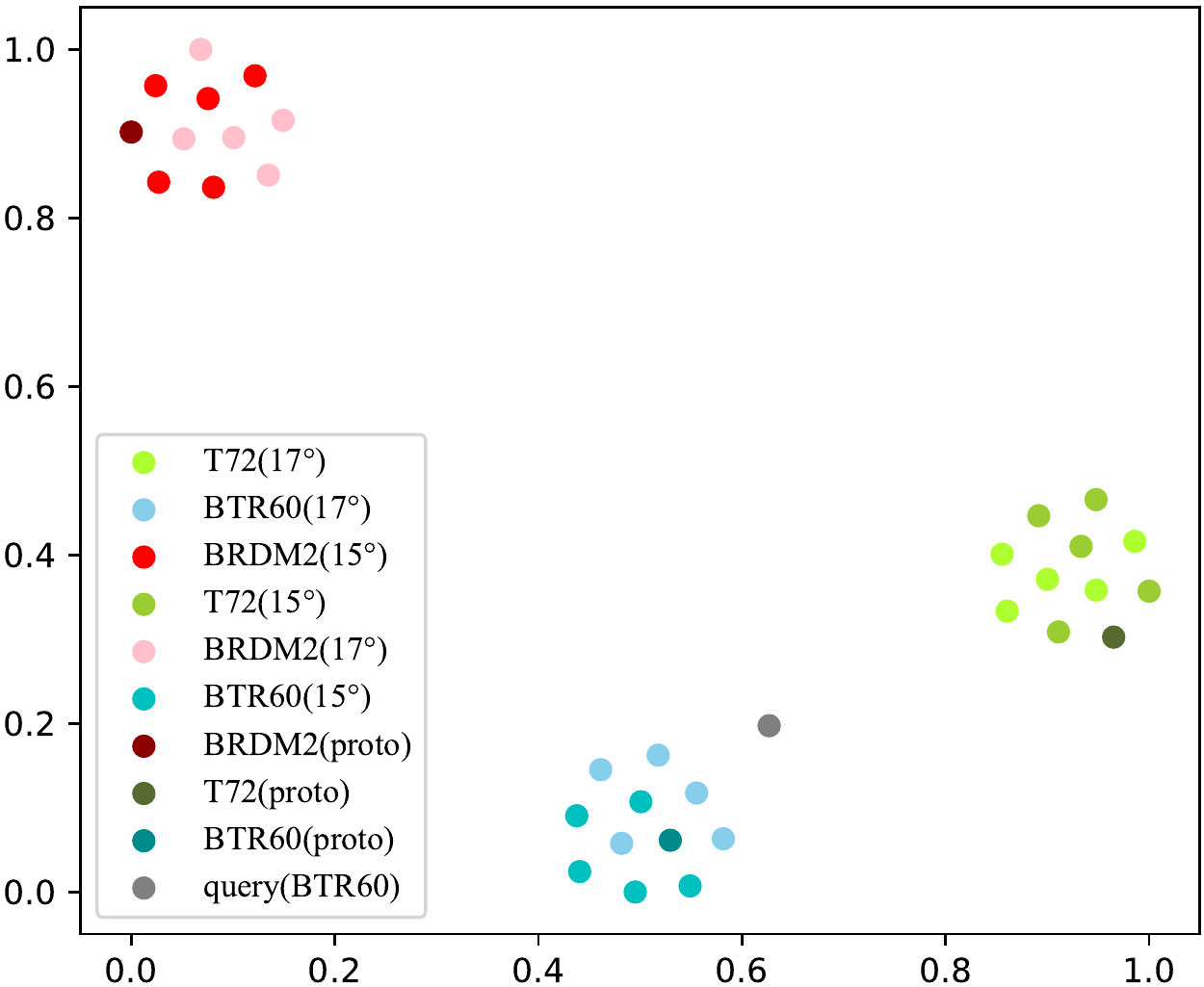}}
\caption{In a 3-way 10-shot task, t-SNE visualization of the distribution of (a) embedding vectors (b) feature vectors after the DGP-Net with non-dense connection (c) feature vectors after the DGP-Net without pro-point (d) feature vectors after the DGP-Net.}
\label{fig_7}
\end{figure}
\begin{table}
\begin{center}
\caption{Classification Accuracies (\%) With Different Ablation Methods in the 3-way and 5-way K-shot Cases on Hybrid-SAR.}
\label{tab1}
\begin{tabular}{ c c c c c c c }
\toprule[0.7pt]
\makecell[c]{\textbf{Methods}} & \makecell[c]{\textbf{3w} \\ \textbf{1s}} & \makecell[c]{\textbf{3w} \\ \textbf{5s}} & \makecell[c]{\textbf{3w} \\ \textbf{10s}} & \makecell[c]{\textbf{5w} \\ \textbf{1s}} & \makecell[c]{\textbf{5w} \\ \textbf{5s}} & \makecell[c]{\textbf{5w} \\ \textbf{10s}} \\
\midrule[0.5pt]
\makecell[c]{GCN} & 78.2 & 84.2 & 88.0 & 20.3 & 20.1 & 20.5 \\
\makecell[c]{GCN(dense)} & 83.5 & 85.3 & 88.2 & 61.2 & 65.3 & 69.6 \\
\makecell[c]{pro-point(only)} & 83.5 & 85.9 & 85.9 & 61.2 & 67.9 & 67.5 \\
\makecell[c]{GCN(dense)+pro\_infor} & 82.9 & 89.4 & 83.9 & 61.0 & 65.5 & 19.8 \\
\midrule[0.5pt]
\textbf{GCN(dense)+pro-point} & \textbf{92.2} & \textbf{93.7} & \textbf{94.2} & \textbf{68.6} & \textbf{76.8} & \textbf{77.0} \\
\bottomrule[0.7pt]
\end{tabular}
\end{center}
\end{table}

\textit{2) Influence of dense connection:} As shown in Table I, if we use the GCN model with non-dense connection, the accuracy of the 3-way task is 78.2\%, 84.2\%, and 88.0\% for 1-shot, 5-shot, and 10-shot. However, for 5-way, the model does not converge. We infer that this is because of the gradient disappearance, and the deterministic similarity information is neglected in the process of learning potential information. If GCN is changed to be densely connected, this problem can be avoided. For the 3-way task, there is a 5.3\% and 1.1\% improvement on 1-shot and 5-shot, respectively. And the accuracy of the 5-way task is 61.2\%, 65.3\%, and 69.6\% for 1-shot, 5-shot, and 10-shot.

\textit{3) Influence of pro-point:} If we only put the prototype nodes into GCN (i.e., input $N$ nodes) after calculating the prototype features, which is named "pro-point(only)" in Table I, the results are better for the small number of samples in $T_{tr}$, but worse for slightly larger number. In this case, the graph network is better at learning relations. The effect is best if we put pro-point into GCN along with other features, which is named "GCN+pro-point". Compared with "GCN(dense)", the accuracies have been greatly improved. We also try to add prototype information into GCN as partial features of nodes (i.e., the length of the input vector is 128), and find that the effect was poor in the case of a large number of samples (such as 10-shot). This method is named "GCN(dense)+pro\_infor". We speculate that it makes half of the feature (prototype part) of the samples in $T_{tr}$ identical, and the whole feature of the sample in $T_{te}$ is different from them. When there are too many samples in $T_{tr}$, the network cannot learn the relations between the prototype part of the sample in $T_{te}$ and that part of the samples in $T_{tr}$. Since the proportion of its influence is $1/2$, the model cannot learn effectively.
\begin{figure}[!t]
\centering
\subfloat[]{\includegraphics[width=0.85in]{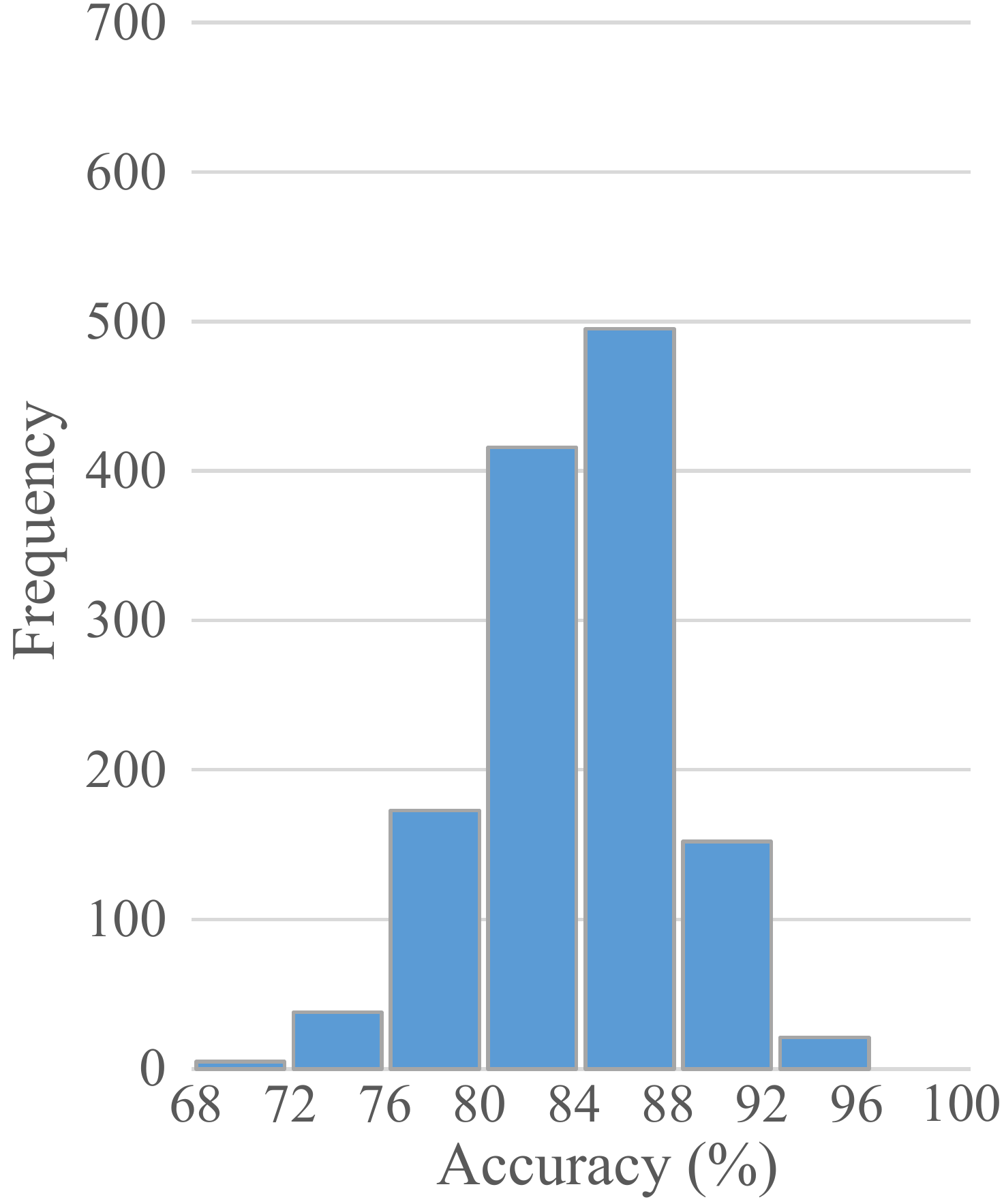}}
\hfil
\subfloat[]{\includegraphics[width=0.85in]{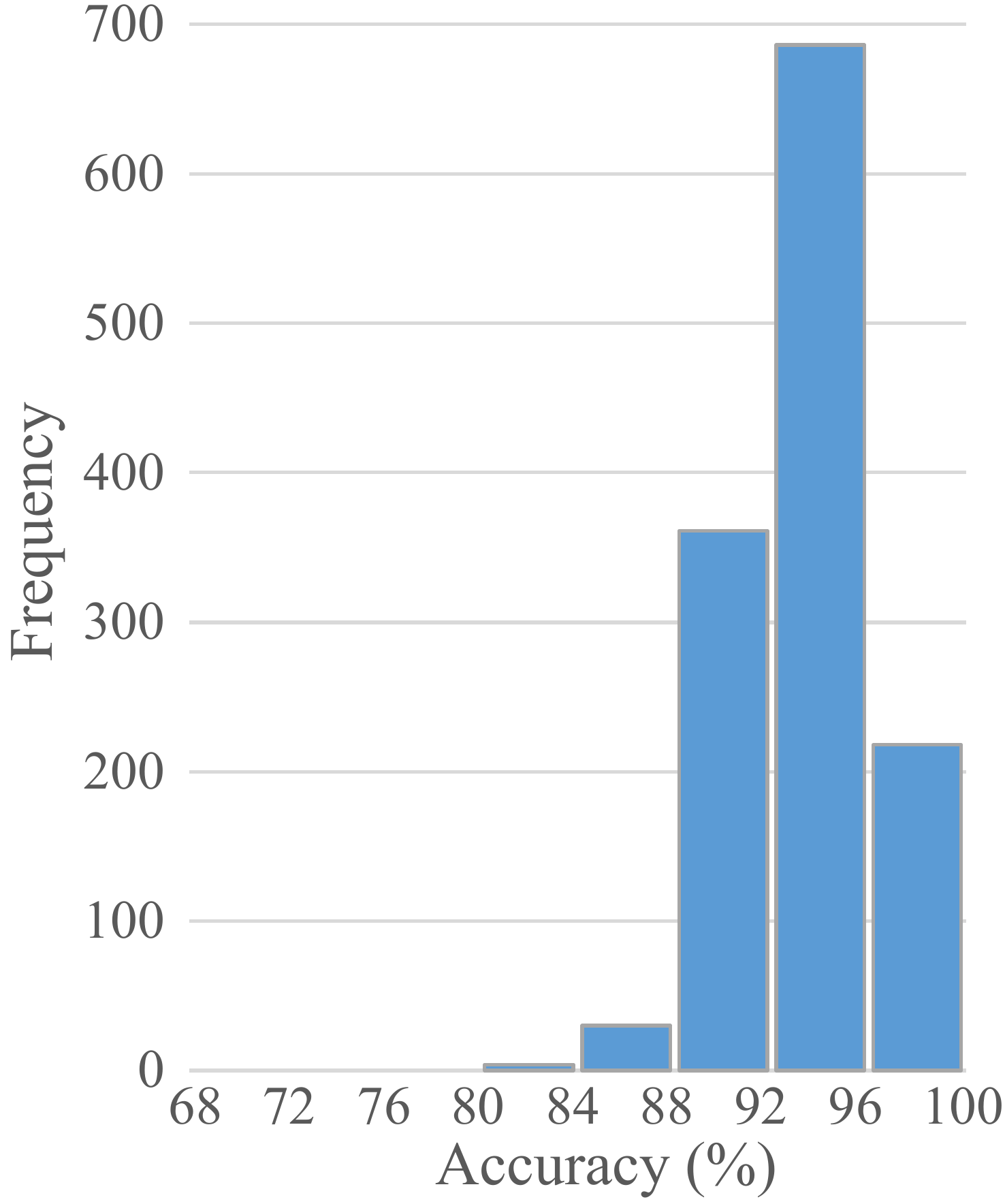}}
\hfil
\subfloat[]{\includegraphics[width=0.85in]{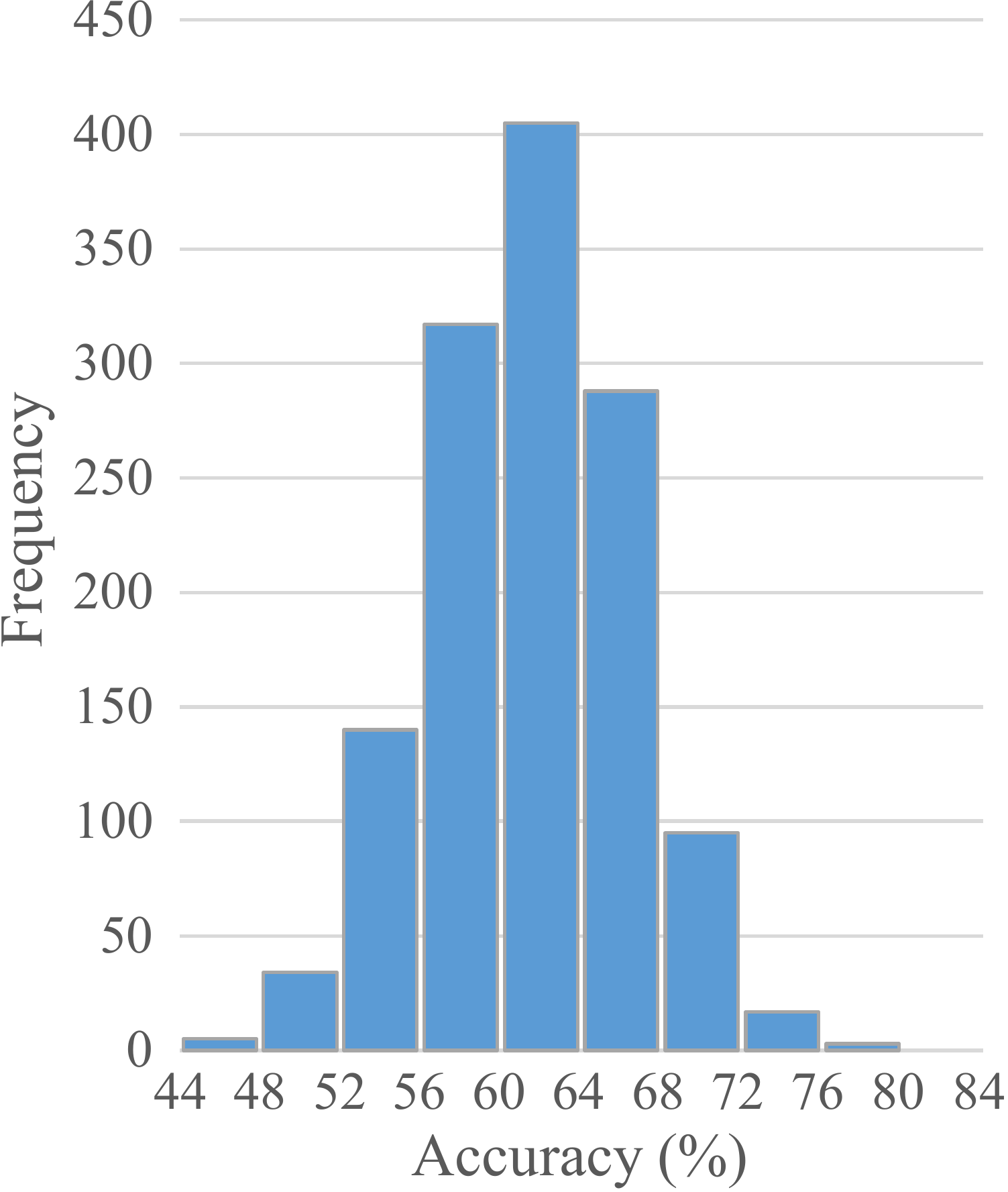}}
\hfil
\subfloat[]{\includegraphics[width=0.85in]{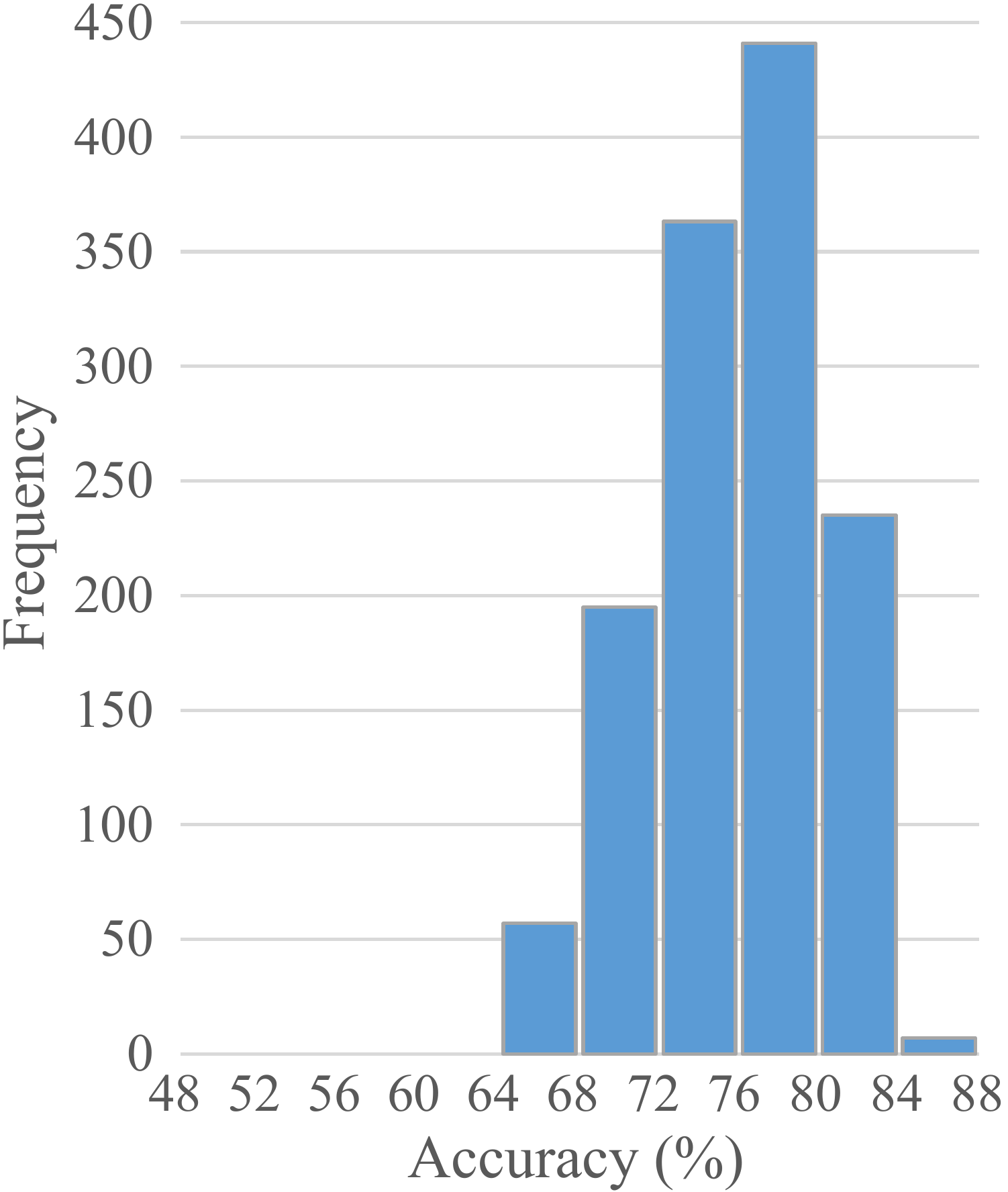}}
\caption{Histograms of the test accuracy (\%) of (a) the DGP-Net without pro-point for 3-way 5-shot (b) the DGP-Net for 3-way 5-shot (c) the DGP-Net without pro-point for 5-way 5-shot (d) the DGP-Net for 5-way 5-shot.}
\label{fig_6}
\end{figure}

For the DGP-Net with or without the pro-point, we respectively construct 1300 independent test experiments through random sampling of 100 samples in each experiment on the Hybrid-SAR dataset. And Fig. 7 presents the histograms of the test accuracy. It can be seen that with the pro-point, the low accuracy due to the sampling problem disappears and the variance $\sigma$ of the accuracy distribution is smaller. Thus, the pro-point does improve the robustness of the model.
\subsection{Performance and Analysis}
We compare our method with several other existing methods (Prototypical Network, Relation Network, Transductive Propagation Network\cite{ref16}, and MSAR\cite{ref15}). As shown in Table II, TPN using GNN has higher accuracy than other typical methods in most cases, which shows that the graph structure model is reliable in FSL of SAR. And DGP-Net is most suitable for the problem setting proposed on SAR and has better classification performance than others in various cases, especially in the case of 5-shot and 10-shot. Because the graph structure model needs the support of a few samples rather than a single sample to learn feature distribution.
\begin{table}
\begin{center}
\caption{Classification Accuracies (\%) With Different Methods in the 3-way and 5-way K-shot Cases on Hybrid-SAR dataset.}
\label{tab1}
\begin{tabular}{ c c c c c c c }
\toprule[0.7pt]
\makecell[c]{\textbf{Method}} & \makecell[c]{\textbf{3w} \\ \textbf{1s}} & \makecell[c]{\textbf{3w} \\ \textbf{5s}} & \makecell[c]{\textbf{3w} \\ \textbf{10s}} & \makecell[c]{\textbf{5w} \\ \textbf{1s}} & \makecell[c]{\textbf{5w} \\ \textbf{5s}} & \makecell[c]{\textbf{5w} \\ \textbf{10s}} \\
\midrule[0.5pt]
\makecell[c]{ProtoNet\cite{ref5}} & 60.6 & 66.6 & 72.3 & 52.5 & 67.8 & 69.9 \\
\makecell[c]{RelationNet\cite{ref7}} & 76.2 & 85.1 & 87.2 & 67.4 & 70.5 & 72.2 \\
\makecell[c]{TPN\cite{ref16}} & 82.2 & 88.3 & 85.7 & 63.8 & 68.7 & 66.3 \\
\makecell[c]{MSAR\cite{ref15}} & 70.8 & 84.7 & 82.1 & 53.6 & 66.8 & 64.7 \\
\midrule[0.5pt]
\textbf{DGP-Net(Ours)} & \textbf{92.2} & \textbf{93.7} & \textbf{94.2} & \textbf{68.6} & \textbf{76.8} & \textbf{77.0} \\
\bottomrule[0.7pt]
\end{tabular}
\end{center}
\end{table}
\section{Conclusion}
In this letter, the DGP-Net framework is designed for the few-shot SAR target recognition to solve the feature deviation problem. In essence, it learns the distribution of features instead of the representation of features, which is more suitable for SAR. The dense network effectively learns potential features, eliminating feature deviation. The prototype anchor points in the DGP-Net gather congeneric samples to improve the robustness of the model. From the perspective of describing the relations between samples, our study confirms that the method based on the GNN family is indeed effective for SAR images. And we need to pay more attention to this in future studies.

\end{document}